\documentclass[conference]{IEEEtran}
\IEEEoverridecommandlockouts

\usepackage[T1]{fontenc}
\usepackage{amsmath,amsfonts}
\usepackage{graphicx}
\usepackage{algorithm}
\usepackage{algpseudocode}
\usepackage{wrapfig}
\usepackage{xcolor}
\usepackage{hyperref}
\usepackage{listings}
\usepackage[capitalise,nameinlink]{cleveref}
\usepackage{subcaption}

\def\BibTeX{{\rm B\kern-.05em{\sc i\kern-.025em b}\kern-.08em
    T\kern-.1667em\lower.7ex\hbox{E}\kern-.125emX}}

\begin{document}

\newcommand{\red}[1]{\textcolor{red}{#1}}
\title{Digital Twin Enabled Runtime Verification for  Autonomous Mobile Robots under Uncertainty}

\author{\IEEEauthorblockN{Joakim Schack Betzer, Jalil Boudjadar, Mirgita Frasheri, Prasad Talasila}
\IEEEauthorblockA{\textit{Department of Electrical and Computer Engineering} \\
\textit{Aarhus University}\\
Denmark }}

\maketitle

\begin{abstract}
 As autonomous robots increasingly navigate complex and unpredictable environments, ensuring their reliable behavior under uncertainty becomes a critical challenge. This paper introduces a digital twin-based runtime verification for an autonomous mobile robot to mitigate the impact posed by uncertainty in the deployment environment. The safety and performance properties are specified and synthesized as runtime monitors using TeSSLa. The integration of the executable digital twin, via the MQTT protocol, enables continuous monitoring and validation of the robot's behavior in real-time. We explore the sources of uncertainties, including sensor noise and environment variations, and analyze their impact on the robot safety and performance. Equipped with high computation resources, the cloud-located digital twin serves as a watch-dog model to estimate the actual state, check the consistency of the robot's actuations and intervene to override such actuations if a safety or performance property is about to be violated. %The runtime verification framework enhances the robot's adaptability to uncertainty and facilitates the detection of anomalies and deviations. 
The experimental analysis demonstrated high efficiency of the proposed approach in ensuring the reliability and robustness of the autonomous robot behavior in uncertain environments and securing high alignment between the actual and expected speeds where the difference is reduced by up to 41\% compared to the default robot navigation control. 
%The findings contribute to advancing the field of runtime verification and autonomous robotics, offering insights into mitigating uncertainties for improved real-time decision-making and performance}.

\end{abstract}


\section{Introduction}
\label{sec:introduction}

Autonomous robots represent a transformative technology that revolutionized various industries and applications in our daily life such as access to hazardous and unsafe environments \cite{JAVAID2021,abbadi2018,Gonzalez2021,LOGANATHAN2023,zhu2020}. Such robots can perform complex tasks with precision and reliability, reducing human intervention and minimizing the risk of errors. An autonomous robot relies on a integrated control system to deliver the expected functionality, by sampling the deployment environment via sensors, analyzing data to estimate the environment state, and compute and execute optimal actuations following the actual state and robot mission \cite{Lewis2018}. Actuation computation is programmed at design stage for a range of known states and configurations, where the robot will be able to autonomously operate and maintain the mission requirements (functionality, performance, safety) \cite{Niloy2021}. However, changes in the deployment environment can lead to uncertain conditions (to be considered as unknown states for the robot). Uncertainty arises from numerous sources, including sensor noise, environment variability and incomplete or inaccurate information about the robot's surroundings \cite{Dreissig23,Tramsen2021}. This uncertainty reduces the robot's ability to recognize its environment state accurately, leading thus to suboptimal actuation or unsafe behavior \cite{De2019}.

Self-adaptivity has been introduced to enable autonomous robots to operate effectively and safely in dynamic and uncertain environments \cite{Hernandez2018,zhu2020}. It relies on adaptive control strategies to learn and reason about uncertainty \cite{Cuevas2019}, but this has many barriers in various robotic applications due to standardization requirements where changes to the robot functionality and performance via self-adaptivity may require a new approval, lack of data related to uncertainty cases, high complexity and computation cost to run self-adaptive algorithms \cite{Cuevas2019,Yang2023}. 

Runtime verification amounts to having an observer (software or model) to monitor and validate the conformity of systems behavior with respect to a set of functional and non-functional properties, written in a temporal logic or as a state machine, in real-time \cite{KRISTOFFERSEN2003}. However, executing runtime monitors on the robot is challenging given that robots are usually equipped with limited computation resources \cite{Bartocci2013}. One way of tackling the challenge of runtime verification and mitigation of uncertainty for autonomous robots is via the use of digital twins (DTs) \cite{Dobaj2022,Temperekidis2022}.   

DTs are having an emerging role in robotics where, as high fidelity executable models to physical systems, they operate via synchronization with the physical system data, actuations and environment \cite{Fitzgerald22}. Thus, performing runtime verification for an autonomous  robot on a cloud-located DT enables on-the-fly monitoring, analysis, mitigation and validation of the robot actuations with respect to actual state,  mission requirements and environment uncertainty \cite{Hoebert2019}.     

This paper proposes a digital twin-based runtime monitoring and verification of an autonomous mobile robot behavior under different uncertainty sources, enabling simulation, real-time monitoring, validation and runtime correction of the robot state with respect to safety and performance requirements. Validation and efficiency analysis are conducted using real-world experiments. 

The rest of the paper is structured as follows: Section \ref{sec:robot-uncertainty} describes the robot and uncertainty sources we consider. Section \ref{sec:related-work} cites relevant related work. The architecture, behavior and operation of the proposed digital twin are specified in Section \ref{sec:digital-twin}. Section \ref{sec:runtime-verification} presents the proposed runtime monitors and verification. Analysis and results are provided in Section \ref{sec:validation}. Finally, Section \ref{sec:conclusion} concludes the paper.

\section{Turtlebot Robot and Uncertainty}
\label{sec:robot-uncertainty}
This section describes the functionality of Turlebot robots and the uncertainties we consider in this paper.

\subsection{Turtlebot3 Burger Robots}
Turtlebotbot3 Burger (\href{https://emanual.robotis.com/docs/en/platform/turtlebot3/overview/}{T3B}) is a highly versatile autonomous (mobile) robot developed by ROBOTIS in collaboration with Open Robotics, an open-source platform that runs on the Robot Operating System (ROS), providing a robust framework for developing and testing robotic applications. It is a two-wheeled differential drive robot empowered with a set of capabilities to deliver a range of missions autonomously (such as reaching a destination, obstacle avoidance, localizing tags, etc) in variable environments and can be precisely controlled with velocity, torque and position \cite{Wan18}.

T3B is battery-powered and relies on a Light Detection and Ranging sensor, called \textit{Lidar}, to sense the environment; a control system deployed on Raspberry PI to analyze data and performance and compute runtime actuations; and 2 wheel motors to  navigate the robot. Using Lidar, the T3B reads a 360 degrees scan of the robot surroundings ($l_1, ..,l_{360}$) to identify obstacles, by measuring the distance on each sampling angle, and building a map of the robot environment \cite{weigl1993}. 

A robot actuation is given in terms of the expected driving speed, composed of linear ($e_l$) and angular ($e_a$) velocity. The expected driving speed is the wheels rotation speed produced by the motors following an actuation command from the control system. T3B turns left and right by trading-off the linear and angular velocity where the higher the angular velocity is the sharper the turn will be. 

Furthermore, we define the actual speed ($a_l, a_a$) to be the speed at which T3B is physically moving at a specific time point. In fact, actual and expected speeds  can differ due to a variety of factors such as floor friction level, ground density conditions (e.g. mud, sand, etc) \cite{dugarjav2013}.

Maintaining an alignment of the expect and actual robot states is of high interest to fulfill the desired performance criteria and safety requirements \cite{LOGANATHAN2023,zhong2021}. For example, when the robot is stuck in low density ground (such as mud or sand) the expected speed can be exponential to the actual speed which may lead to drain the robot battery and violate the mission delivery time \cite{Luan20}. The complexity of the problem can become intractable, especially when navigating through challenging scenarios, such as dynamic environments \cite{abbadi2018}. Beside the sophisticated control algorithms \cite{erickson2013}, combining runtime monitoring and mitigation \cite{Carmelo20,Petrovska21} is one of the different techniques proposed in the literature to cope with inaccurate state estimation for autonomous mobile robots and divergence of the expected and actual performance \cite{zhong2021}. 

\subsection{Uncertainty Sources}
Autonomous robots increasingly navigate complex and unpredictable environments, where operation conditions may vary to the scale that is beyond the configurations such robots are engineered for  \cite{Laxman20}. These dynamic contexts are often accompanied by the pervasive presence of uncertainty, that poses many challenges to ensuring reliable behavior and high performance of autonomous robots \cite{Andersson2020}.

The uncertainty sources we consider in this paper are the following:
\begin{itemize}
\item Faulty Lidar readings due to noise, dust (or any airborne particulates) or other obscurant media in the deployment environment, leading thus to erroneous data, incomplete or inconsistent image of the robot environment \cite{Dreissig23}.
\item Dynamic environment conditions related to variable floor friction and density due to presence of mud, sand or highly lubricated surfaces that degrade the friction level and the locomotion efficiency of the robot maneuvers \cite{Tramsen2021}. 
\end{itemize}

Uncertainty impacts the robot decision making in a way that the environment state cannot be recognized upon which proper actuations can be defined efficiently. Moreover, it is by far not possible to examine the sources and manifestations of uncertainty and their impact on the robot at design stage \cite{Gonzalez2021}. Thus, runtime mechanisms are needed to accommodate the rule-based control of autonomous robots to capture, understand and mitigate uncertainties \cite{zhu2020}.

To capture the uncertainty related to faulty Lidar readings, we propose thorough data analysis and proper correction actions at runtime. Since those operations are computationally expensive to execute in T3B, we develop runtime monitors, deployed as cloud-located digital twins, to investigate the Lidar data consistency. Moreover, we augment T3B with Simultaneous Localization and Mapping (SLAM) \cite{temeltas2008} and couple it with digital twins to achieve runtime monitoring and mitigation of the uncertainty related to dynamic floor friction and density. In fact, SLAM enables to estimate the actual robot speed that is highly dependent on the floor friction, density and collisions. The runtime monitors enable to impose constraints on the robot behavior to mitigate the uncertainty impact and maintain high performance at runtime, e.g. actual speed must not differ from expected speed with certain threshold.

\section{Related Work}
%\red{Gita -- not done yet with this}
\label{sec:related-work}

%A-Overview of Runtime Verification in Robotics B. Challenges Posed by Uncertainty in Autonomous Robot Behavior C. Existing Approaches for Handling Uncertainty in Robotics D. Role of Digital Twins in Enhancing Reliability and Adaptability

Autonomous robots are expected to reduce human efforts in a variety of domains, from replacing heavy manual labor (e.g. in agriculture), to being deployed in remote and harsh environments dangerous to humans (e.g. de-mining missions, search and rescue).
While there are examples of such robots already available in the market, the reader is pointed to the Robotti field robot~\cite{foldager2018design}, their capabilities are still quite limited in the face of unforeseen circumstances.
This is due to the real world environments being unstructured, partially or fully unknown, and possibly uncontrollable.
To be able to fulfil their missions successfully and safely, autonomous robots have to adapt to unforeseen issues and events~\cite{lee2019robust, ahmadi2016robust}.

Different factors can be a source of uncertainty: faulty components, noisy sensors and actuators, malicious attacks, uncertainties in the environment itself, accrued model errors~\cite{zhu2020know}. 
Designing good models is not a trivial task due to the complexity of the involved software and hardware components, environment, and the resource and computational limitations of the embedded devices typically used in robotics. Additionally, as revealed by a recent survey, planning seems decoupled from sensors such as lidars and cameras, which if incorporated properly could help dealing with uncertainty~\cite{LOGANATHAN2023}.

Reinforcement learning has sparked interest in the robotics community~\cite{singh2022reinforcement}, for manipulators, trajectory tracking, navigation and path planning~\cite{zhang2021reinforcement}. 
While great success was shown in simulation, applicability in the real world remains challenging~\cite{dulac2019challenges,smyrnakis2020multi}.
This is a result of the limited number of uncertainty samples that can be used for learning, high training costs and uncertain models \cite{ahmadi2016robust,zhang2021reinforcement}.
In addition, it is necessary to provide real-time inference, while simultaneously dealing with large or unknown delays in sensors, actuators, and rewards~\cite{dulac2019challenges}.

DT technologies have garnered quite the traction in recent years, as they promise to optimise the development and deployment of robots, and cyber-physical systems (CPSs) in general, enabling services like runtime monitoring, adaptation, and system reconfiguration to adapt to changes in the CPSs themselves as well as their environment~\cite{feng2021introduction}. 
The DT could be run in the cloud, thus alleviating problems regarding resource and computational limitations, and extending the capabilities of its physical counterpart. 
Dobaj \textit{et al.} take a DevOps approach in order to support adaptations and the verification thereof at the CPS level~\cite{Dobaj2022}. 
Specifically, run-time verification is performed to ensure that a service B, that is to replace an existing service A, provides reasonable output before it can affect the CPS. 
Allamaa \textit{et al.} have proposed an approach to adapt controllers created in simulation to a real system (e.g. an ECU)~\cite{allamaa2022sim2real}, by effectively transferring control parameters estimated in simulation to the real system while taking into account noise and edge cases. 
In other scenarios, such as those including Cyber-physical Production Systems (CPPSs), the DT can be adopted to monitor and generate new strategies to optimize the production process when new orders are issued~\cite{kang2019design}. 
The results from the DT deliberation are thereafter verified and validated regarding which actuation strategy to apply to actuate the system. Rivera \textit{et al.} propose a DT-based reference architecture for the development of what they call smart CPSs (SCPSs)~\cite{rivera2021designing}, where an efficient actuation is derived from the multiple outcomes of multi-DT systems for CPSs. 
Additionally, they adopt the concept of viability zones, where reference signals used in the DT are updated as the divergence to the corresponding real signals increases. 

%DIGITAL TWIN AND HMI FOR COLLABORATIVE AUTONOMOUS MOBILE ROBOTS IN FLEXIBLE LOGISTICS --> phd thesis

%Cloud-Based Digital Twin for Industrial Robotics ---> this seems a conceptual solution, but they say in the paper they applied to a real case but I don't see any actual results/evidence.

\section{Digital Twin Specification}
%\red{Prasad}
\label{sec:digital-twin}

\begin{figure*}
\centering
\begin{subfigure}[t]{0.6\textwidth}
    \includegraphics[width=\textwidth]{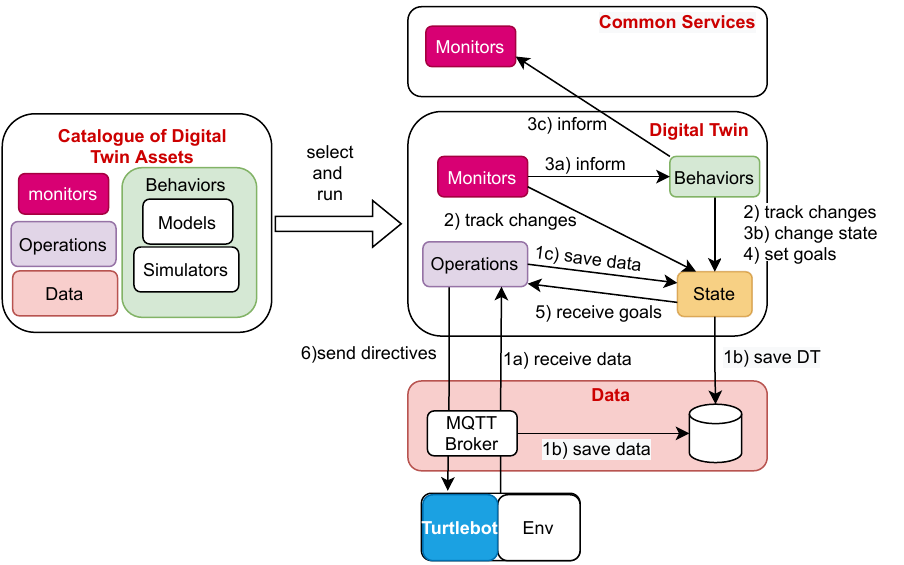}
    \caption{DT Architecture. The sequence of interactions leading to monitoring results are labeled with numbers.}
    \label{fig:dt-architecture}
\end{subfigure}
\begin{subfigure}[t]{0.2\textwidth}
    \includegraphics[width=\textwidth]{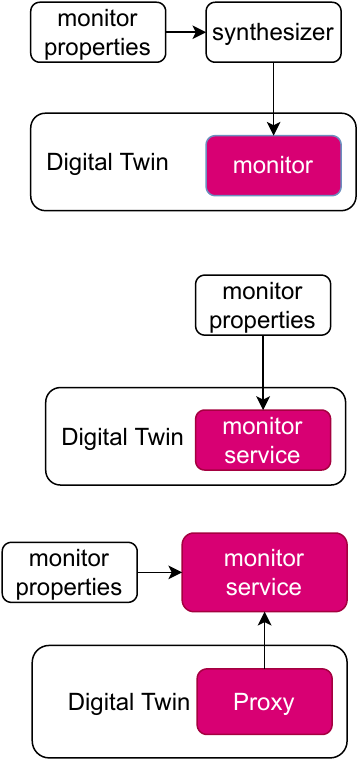}
    \caption{Different possibilities in integration of monitors into the DT}
    \label{fig:monitors-integration}
\end{subfigure}

\caption{Digital Twin architecture and monitor integration.}
\label{fig:dt-monitors}
\end{figure*}

A DT has been widely defined as a digital representation of a physical object, that through data exchange, reflects the evolution of the physical twin (PT) over time and in turn influences the future behaviour of the PT~\cite{Grieves17}. This definition highlights the bidirectional data exchange over a communication infrastructure and the linked evolution of DT vis-a-vis PT. In addition, the need for multi-domain, multi-scale modeling approaches became apparent in the creation and evolution of DTs~\cite{mckee2023platform,Zambrano22}. The simulators are required for models execution. A pair consisting of a model and simulator gives raise to behaviors which simulate one aspect (part in common terms) of existing or desirable behavior of the PT~\cite{liu2021review}. It is often desirable to have a catalogue of suitable \textit{models, simulators} and \textit{behaviors} to be selected and used in the DT.
Safety is a must in autonomous systems and monitors play significant role in maintaining it \cite{feng2021introduction,lee2019robust}.  %Monitors are programmable and are configured with safety properties. 
%The runtime monitors are more appropriate in autonomous robotic systems like T3B. 
Both monitors and their safety properties are general and thus belong to a catalogue of DT assets.

In addition, the \textit{data} whether live or historic significantly impacts the performance of DT in fulfilling its obligations to PT. It is often the case that the data itself is collected and reused, and can either come from DT or its environment. Such data is helpful in planning the optimal behavior for PTs like autonomous robots. Thus, the data itself can be put into a catalogue of assets from which many DTs can be built~\cite{Talasila2023}. 

Given that PTs like Turtlebot supports many protocols - MQTT, RabbitMQ, HTTP etc. - it is often desirable to have implementations of the data interfaces in the catalogue itself under a form of linked data. Moreover, turtlebot robots have their well-defined Application Programming Interface (API) over which the remote software including DT can interact with PTs. The software packages providing access to these APIs are often needed but are not integral to the DT itself. In addition, it is often necessary to operate at an abstract level, e.g,. goal versus concrete instructions. The \textit{operations} provide a way to convert these DT-level goals to concrete PT-level instructions.

Figure~\ref{fig:dt-architecture} shows an architecture of the DT we propose. The catalogue of DT assets from which the DTs are created is highlighted. The DTs created are templates (classes) from which many DT instances (objects) can be created. In this paper, the discussion is oriented towards one DT instance for T3B and this instance is referred to as DT itself. Figure~\ref{fig:dt-architecture} also shows the sequence of operations taking place between T3B and its DT. One cycle of operations is enumerated in this figure. The data is collected from both T3B and its environment and communicated to DT. The collected data contains the relevant state of PT. In the present case, the expected linear and angular velocity, actual speed, Lidar data and the current actuation. %The data can also contain the LiDAR and any other readings of the environment.
The data is sent over the MQTT broker to two end points, namely DT~(\textit{step-1a}) and data storage~(\textit{step-1b}). Both the MQTT broker and data storage constitute the data part of the DT. The data is received by the operations interface of the DT (using the Telegraf Connector to be explained in Section \ref{sec:runtime-verification}.C) and becomes a part of DT state~(\textit{step-1c}). It is pertinent to note that the data (there by state) received from the PT is only a subset of the DT state which contains the state derived from behaviors. The changes in the DT state triggers both monitors and behaviors~(\textit{step-2}). The monitors check for compliance with safety properties while behaviors simulate models with the existing state as input. The monitors can potentially influence simulations by communicating the safety compliance status of the PT~(\textit{step-3a}). After conclusions of (potentially multiple) simulations, the behaviors change the DT state and set new goals for PT~(\textit{steps-3b and 4}). The operations then convert these goals to concrete directives and send them to PT~(\textit{step-5}).

Digital Twin platforms supporting and executing many DTs often provide a service layer on top of DTs. The service layer is fed by DTs based on their internal states. The users can interact with service layer to perform monitoring, visualisation, analysis, global planning and decision making tasks~\cite{ferko2022architecting,LumerKlabbers&21}. It is also possible to place monitors as a common service on DT platforms as well in which case multiple DTs can use them. 
The communication of other parts of the DT with monitors can follow either push or pull patterns. The pull communication pattern indicates the active checking and copying of the DT state by the monitor. This pattern is most appropriate for monitors integrated into the DT. The push communication pattern indicates the activation of monitors by sending updated state. In this case, the monitor does not need access to DT state and thus this communication pattern is most appropriate for monitors placed in common services~(\textit{step-3c}).

%\subsection{Monitor Integration}
There are multiple patterns in the integration of monitors into DTs. These patterns are illustrated in Figure~\ref{fig:monitors-integration}. %The relevant explanation follows.

\paragraph{Synthesizer} The run-time monitoring tool uses monitor properties to synthesise the monitor which then is used inside DT. A just-in-time synthesis is advantageous over the  of monitor executables. This paper implements a synthesized monitor.

\paragraph{Private Service} The monitor is reusable and is integrated as a service into the DT. This service is exclusively used by one DT. Given the private scope of the monitor, the implementation and integration are less complex.

\paragraph{Public Service} The monitor is placed as a common service and is reusable across many DTs. However, the monitor must have the ability to simultaneously service over many DTs.

%Some tools including TeSSLa used for implementation do take the monitor properties. TeSSLa can either create a synthesized monitor or provider monitor service. 
A common thread across all three integration patterns is the externalisation of monitor properties. These must not be baked into the monitors themselves. It is also advantageous to have the ability to make dynamic changes to the monitored properties of autonomous robotic systems but supporting such a flexibility in monitoring tools is not trivial.

\section{Digital Twin-Based Runtime Verification Framework}
\label{sec:runtime-verification}
The proposed runtime monitors amount at observing the robot and environment state, that could be partial due to uncertainties, and validate the robot runtime actuations with respect to a set of safety and performance properties.

\begin{figure}
\centering
\includegraphics[scale=0.45]{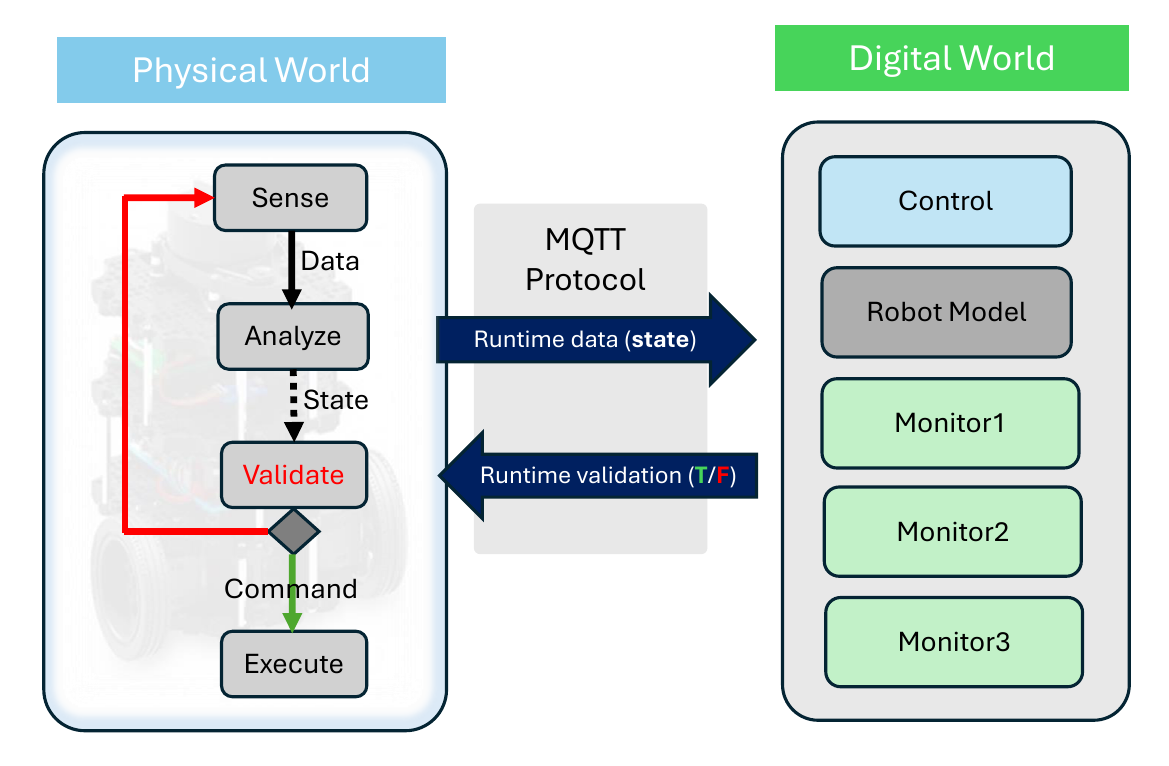}
\caption{Proposed framework for DT-enabled runtime verification of T3B.}
\label{fig:overall}
\end{figure}

Figure~\ref{fig:overall} depicts the overall workflow of the proposed DT-enabled runtime verification for T3B robots under uncertainty. In fact, the robot senses its environment (\texttt{Sense}), analyzes such data in correlation with the actual speed and computes the corresponding actuation command (but does not execute it yet ) (\texttt{Analyze}). 
The state formed by sensor data, computed actuation and both actual and expected velocities is sent to DT via MQTT protocol to enable runtime monitoring and verification of the consistency of the actual actuation command with respect to different properties. Upon receiving the validation outcome (\texttt{Validate}), the robot executes the actuation command if approved by the DT runtime monitors. Otherwise, the robot samples again the environment to estimate a new state and compute new actuations.   

\subsection{Safety and Validation Properties}
Safety is an essential consideration to be mindful of both at design and during runtime as malfunction (safety violation) could result in equipment-related or environmental damages. The safety property we consider aims to prevent robot collisions
by maintaining a braking distance to any obstacles. The braking distance is dependent on the actual speed, and must be larger than the distance to any given obstacle, to a degree that is considered acceptable in a given scenario.
As such, the runtime actuation of T3B must account for the actual speed and the distance to obstacles in the robot trajectory. Formally, such a safety property is specified as follows:

\[ \textbf{P1}:~~  \forall i~Bdist(s_i) \le Ldist(s_i) \]

\noindent Where $s_i$ is a runtime state of T3B, $Ldist(s_i)=min(s_i.l_{330},..,s_i.l_{30})$ is the distance to the obstacles located in the current heading angle of T3B, and $Bdist(s_i)$ is the  distance needed at state $s_i$ to bring the robot to a full stop. In fact, the braking distance depends on the actual speed and computed from linear and angular velocities \cite{Salimi20}.

Due to the environment uncertainty, the likelihood of the actual speed (terrain traversing speed) to deviate from the expected speed (wheels spinning speed) increases. This gap becomes evident when the robot operates in terrains with variable friction and density. 
To prevent drastic divergence, we impose a tolerance property to limit how far actual speed can deviate from expected speed at runtime as follows:

\[ \textbf{P2}:~~\forall i~(s_i.e - s_i.a ) \le \delta \]

\noindent Where $s_i.e$ and $s_i.e$ are the expected and actual speeds respectively, and $\delta$ is the maximum tolerance. Integrating this runtime property ensures that the robot can receive appropriate commands to correct any discrepancies between the actual and expected speeds.

The Lidar can occasionally produce inaccurate and flawed readings,
and imposing a runtime monitor to ensure that the robot does not make misguided navigation decisions due to such faulty readings is of high significance. We specify a validation property for the Lidar readings as follows:

\[ \textbf{P3}:~~\forall i~j, ~(s_i.l_j - s_i.l_{j+1}) \le \gamma \wedge (s_i.l_j - s_i.l_{j-1}) \le \gamma \]

This property enables to detect erroneous readings where the value of a Lidar angle $l_j$ of a given state $s_i$ deviates drastically from both adjacent angles. Such cases are identified as not obstacles given that an obstacle cannot be captured only by a single angle, thus either a dust particle or a faulty reading. 

\subsection{Synthesis of Runtime Monitors}
To impose the satisfaction of the aforementioned properties at runtime, we synthesize a runtime monitor for each property using the {TeSSLa} tool \cite{Kallwies22}. In fact, TeSSLa is a framework that enables instrumentation and streaming of programs execution using observers for runtime verification purposes. Endowed with back-end synthesizers \cite{Havelund2002} and a friendly instrumentation language, TeSSLa synthesizes monitors for behavior and property specifications.    

\begin{figure}
    \centering
    \includegraphics[scale=0.23]{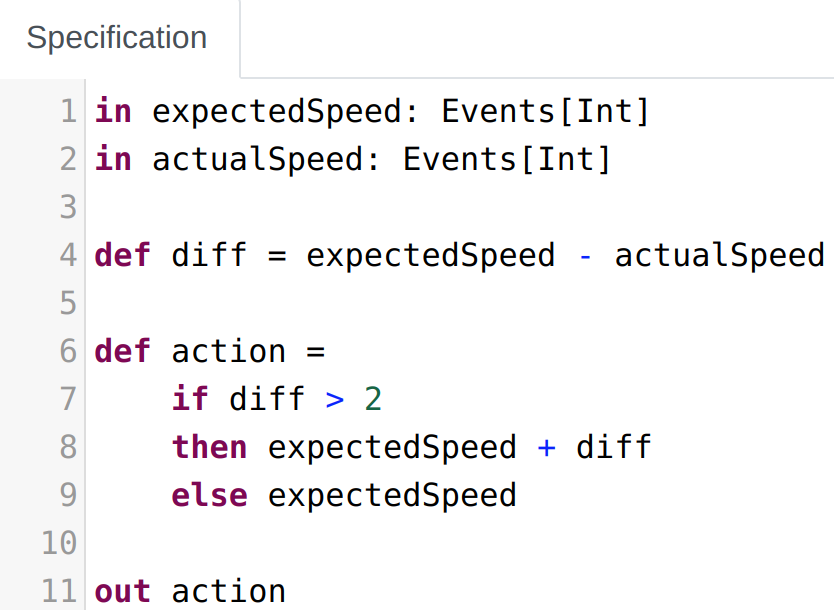}
    \vspace{-1mm}
    \caption{TeSSLa monitor specification for property \textbf{P2}.}
    \label{fig:77}
    \vspace{-3mm}
\end{figure}

Figure \ref{fig:77} depicts the TeSSLa specification of the runtime monitor implementing the tolerance  property \textbf{P2}. The specification takes two input streams, \textit{expectedSpeed} and \textit{actualSpeed} computed from $e_l$, $e_a$, $a_l$, and $a_a$. 
The input streams are used to calculate the speed difference \textit{diff} (could be negative), which forms a baseline  to compute the appropriate actuations. For example, \textit{diff} can be added to \textit{expectedSpeed} to make the actual speed closer to the expected speed. The output from this TeSSLa specification is a Boolean value about whether the expected speed resulting from the execution of the current actuation would violate this property or not, beside to the action representing the adjusted expected speed, which then can be used for further process or monitoring.

The runtime monitors synthesized using TeSSLa ensure that T3B behavior adheres to the specified properties P1, P2 and P3. %In this context, we aim to optimize the actual speed of T3B given the expected speed, subject to the tolerance constraint \textit{P2}. 
For example, the runtime monitor in \ref{fig:77} ensures that at every state, the absolute difference between these two does not exceed the tolerance $\delta$. To illustrate the data flow synthesis and test of the monitor specification, we utilize \href{https://play.tessla.io/}{TeSSLa's playground}. Using actual states data, we create a trace file, serving as a set of input events to the monitor, we can observe the output from the monitor specification as depicted in Listing~\ref{lst:trace_input}. 

\begin{center}
    \begin{lstlisting}[caption={Input Trace to P2 monitor.}, label={lst:trace_input}, basicstyle=\small]
        0: actualSpeed = 0
        0: expectedSpeed = 1
        2: actualSpeed = 5
        2: expectedSpeed = 1
        4: actualSpeed = 2
        4: expectedSpeed = 5
        6: actualSpeed = 3
        6: expectedSpeed = 3
        8: actualSpeed = 1
        8: expectedSpeed = 4
    \end{lstlisting}
\end{center}

\begin{figure}
    \centering
    \includegraphics[scale=0.22]{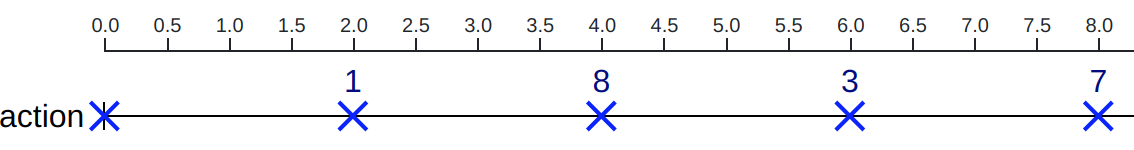}
    \caption{TeSSLa output stream visualization.}
    \label{fig:tesslaOutput}
\end{figure}

The trace file is a sequence of timestamped events that represent the values of the input streams at specific points in time. Processing the events from the actual trace file at runtime produces the output streams shown in Figure \ref{fig:tesslaOutput}.  %The figure visualizes the output stream at the given time stamps.
TeSSLa provides us with the output from processing the events against the monitors specification, effectively testing and validating the robot behavior.

%\cite{Synthesizing Monitors for Safety Properties}
T3B controller relies thereafter on combining the validation outputs from the different runtime monitors together with a simple \href{https://github.com/husarion/static_docs_deprecated/blob/master/src/core2/tutorials/ros-tutorials/3-simple-kinematics-for-mobile-robot.md}{control logic} to decide the runtime actuations. However, in this paper we considered the case where DT runtime monitors only validate whether the newly computed actuations by the robot itself are going to maintain the different properties on the actual state.      

%\red{Jalil} Outcomes from runtime monitors

\subsection{Integration of Runtime Monitors into DT}
The integration amounts at mediating the outcomes of the runtime monitors to the control module, establish communication between the DT and the physical T3B and develop an orchestrator that invokes the runtime monitors execution.   
We use the \href{https://mqtt.org/}{MQTT} protocol to transfer data and the validation outcomes between T3B and the  DT. Furthermore, we utilize the \href{https://docs.influxdata.com/telegraf/v1/}{Telegraf} open-source agent to collect and analyze events from the MQTT streams. In fact, Telegraf  provides the functionality to create pipelines for listening to events from the robot, parse the data through a TeSSLa specification file, and send back the TeSSLa output stream to the robot.

\begin{center}
    \begin{lstlisting}[caption={MQTT listener}, label={lst:MQTTListener}, basicstyle=\small]
[[inputs.mqtt_consumer]]    
 servers = ["tcp://test.mosquitto.org:1883"]
 topics = ["tessla"]
 data_format = "json"
 json_string_fields = ["actualSpeed"]

[[outputs.mqtt]]
 servers = ["tcp://test.mosquitto.org:1883"]
 topic = "action"
 data_format = "json"
    \end{lstlisting}
\end{center}
As illustrated in Listing \ref{lst:MQTTListener}, we specify the topic \textit{tessla} that Telegraf will be listening to, the format of the data, and which JSON string field to look at. We also specify the topic where Telegraf will output the data. The input data is processed and forwarded to specific outputs, which in this case, will be TeSSLa for temporal monitoring and analysis. To be able to do this, Telegraf has created the \href{https://git.tessla.io/telegraf/tessla-telegraf-connector}{TeSSLa Telegraf Connector}, which takes a TeSSLa specification and uses the TeSSLa compiler to convert it into a Rust project. This results in a Rust program that can interact with Telegraf, enabling the seamless integration and real-time processing of data streams based on the specified TeSSLa properties. Furthermore, we include the \textit{Telegraf.tessla} file into the TeSSLa specification as follows:
\begin{center}
\textit{def @TelegrafIn(id: String, tags: String, field:String)}
\textit{def @TelegrafOut(name: String)}
\end{center}

The \textit{TelegrafIn} marks the input streams and \textit{TelegrafOut} marks the output streams, as we previously specified in Listing \ref{lst:MQTTListener}. This is then tied to the events in the specification file for the input streams, and the output for the output streams.

The last configuration step of the TeSSLa Telegraf Connector is to create communication between the generated Rust project and Telegraf. The Rust program  connects to Telegraf via UDP, and based on the incoming data points from Telegraf, the input functions of the TeSSLa monitor are called. 

The output streams from the TeSSLa monitor are translated into InfluxDB format and sent to Telegraf via UDP, providing thus a comprehensive and complete solution for real-time monitoring, analysis, and management of complex data streams.
% the pipeline is complete. By using TeSSLa in conjunction with Telegraf and the Tessla Telegraf Connector, we can leverage the strengths of both systems, providing a comprehensive solution
%T3B times the connection to DT so that if connection lost (too long, received packets corrupted, etc), T3B does not wait forever and rather m,aintain the actuation frequency that is 500 milisecond****

\section{Experiment and Validation}
\label{sec:validation}
To validate the proposed DT-enabled runtime monitoring and verification for T3B under uncertainty, we implemented the DT, runtime monitors, the MQTT protocol and Telegraf interface. We conducted different experiments to demonstrate the efficiency of using runtime monitors. Furthermore, we collected actual (time-stamped) robot data to provide the possibility to run different simulations in the DT without the need to synchronize with the physical robot.

%Figure~\ref{fig:XX} depicts a code snippet of the runtime monitor to observe and validate the deviation of actual speed from expected speed. The monitor is synthesized from the property ** using TeSSla tool ***** 

\subsection{Experiment setup}
The experiment setup involves a T3B running in an uncontrolled environment and  publishes its data through ROS topics. The data is then processed through an MQTT broker, which acts as an intermediary between the ROS system and the DT both ways. From T3B to DT, the broker facilitates communication by subscribing to the ROS topics and publishing the topics data to specific MQTT topics, which the DT then subscribes to. Similarly, the DT publishes data back to a different MQTT topic, which the broker subscribes to, and publishes it to ROS topics which T3B subscribes to. The setup of having the communication between T3B and DT done through MQTT ensures a decoupled architecture, where both ends can operate independently. % of each other's codebase, the turtlebot does not need to include MQTT-specific code, and the DT does not need to handle ROS-specific code.
Additionally, this architecture simplifies the creation of a mock version of the physical robot. By publishing data in the same format as T3B to the ROS topics, the MQTT broker gathers data and forward it to DT, seamlessly integrating the mock data into DT.

\begin{figure}
    \centering
    \includegraphics[width=\linewidth]{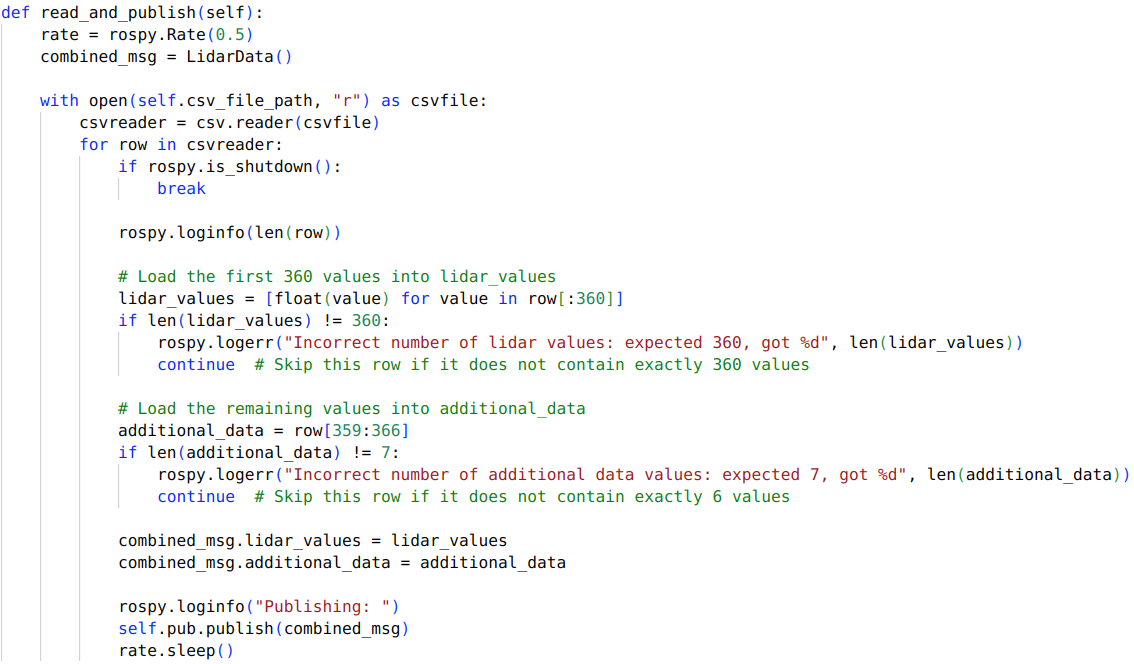}
    \caption{Code snippet of the script to simulate stored data.}
    \label{fig:runSimulationCodeSnippet}
\end{figure}

\begin{figure}
    \centering
    \includegraphics[width=\linewidth]{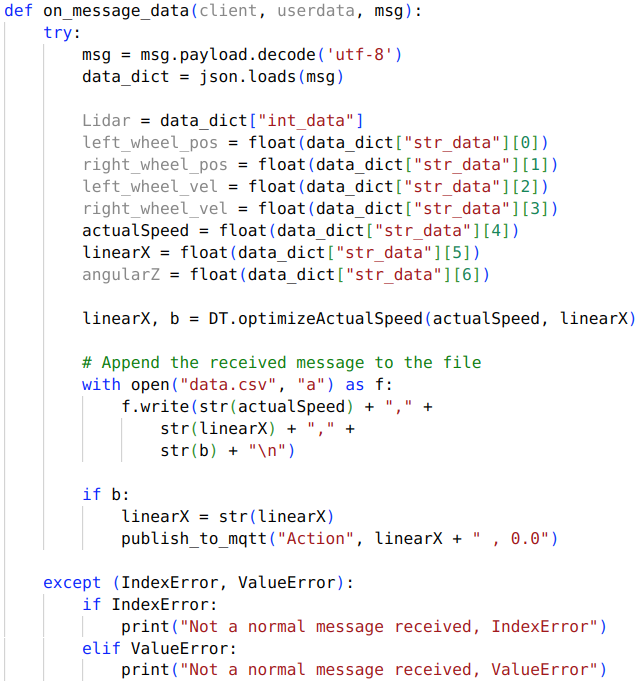}
    \caption{Code snippet of DT to call the tolerance monitor using data.}
    \label{fig:DTCallCodeSnippet}
\end{figure}

Figure \ref{fig:runSimulationCodeSnippet} presents a code snippet demonstrating how to publish mock data from a CSV file to a ROS topic. As long as the data is structured in the Lidar data format, a custom ROS message comprising two arrays (one with 360 float values and the other with 7 string values), it is straightforward to simulate the physical robot in DT. In Figure \ref{fig:DTCallCodeSnippet}, we depict how the arrays from the Lidar data are used by DT .

\begin{figure}
    \centering
    \includegraphics[width=\linewidth]{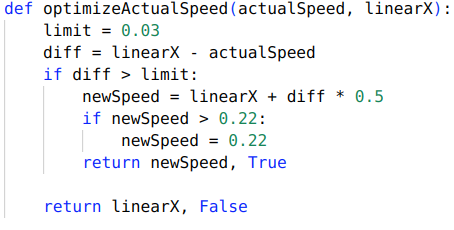}
    \caption{Code snippet of the tolerance monitor implemented in Python.}
    \label{fig:performanceCodeSnippet}
\end{figure}

In this experiment, we focused only on the tolerance runtime monitor (synthesized for P2), which is why only the actual speed and linear speed variables are defined. Furthermore, we enable the runtime monitor to propose corrective actions to the speed actuation, beside to validation. The DT calls the \textit{optimizeActualSpeed} function, depicted in Figure \ref{fig:performanceCodeSnippet}, which is an implementation of the corresponding TeSSLa monitor. 

The speed difference is a proportional gain of 0.5 ensuring that the corrective action taken to adjust the robot's speed is proportional to the difference. This ensures that larger errors result in stronger corrective actions, improving the T3B's ability to reach the desired speed more accurately. Furthermore, we also check that the proposed speed actuation does not exceed 0.22, as this is the maximum speed for T3B. Finally, we are also returning a boolean value, which is used to visualize which data is augmented and to check when we need to publish an action for the robot to execute. %For this experiment, we only focused on the linear speed, as the physical experiment setup involves driving in a straight line. 

\subsection{Use of SLAM methods}
\begin{figure}
    \centering
    \vspace{-3mm}
    \includegraphics[scale=0.31]{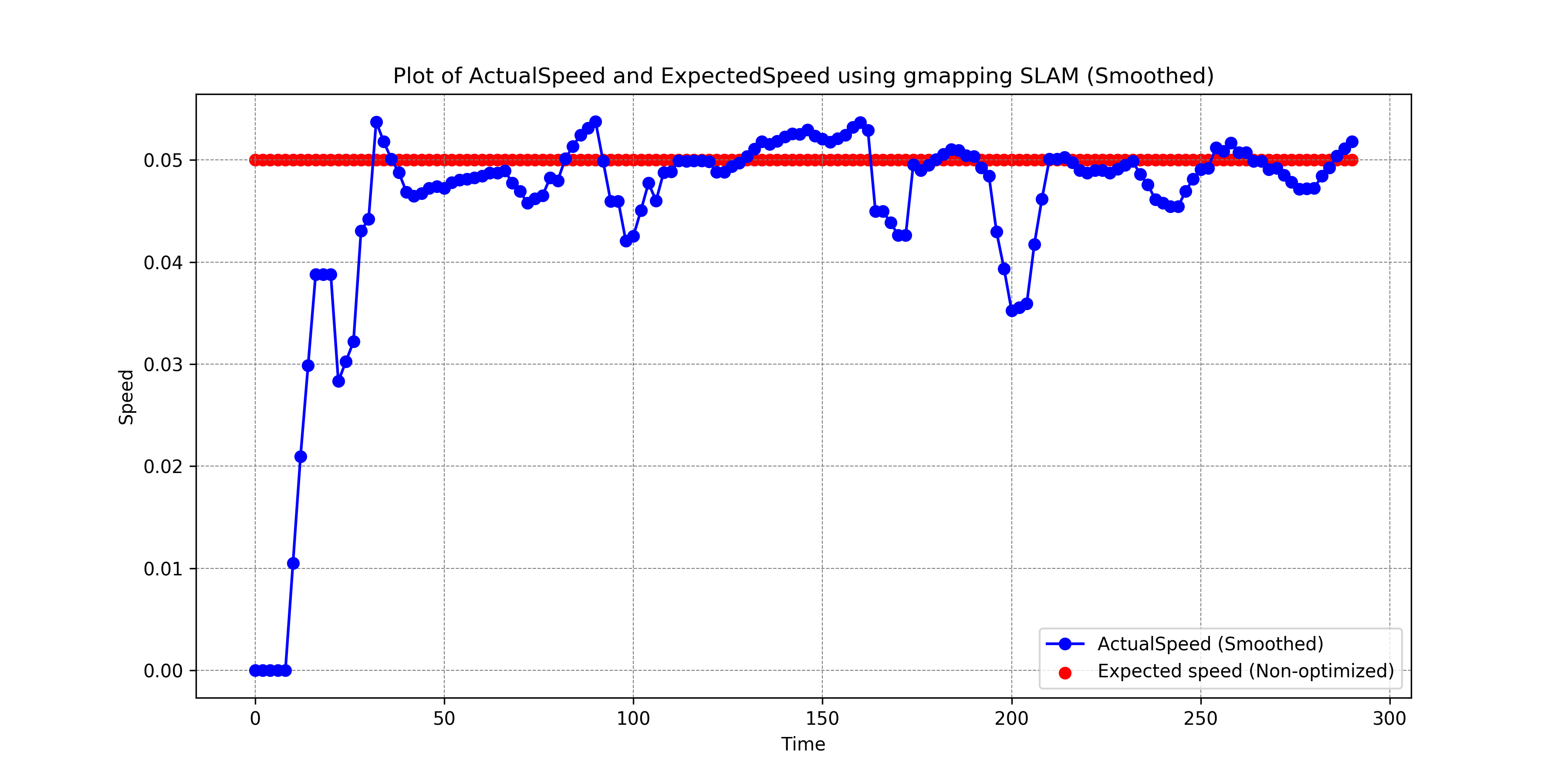}
    \caption{Plots showing a stationary T3B publishing its actual and expected speed using SLAM gmapping (top figure) and SLAM hector (bottom figure).}
    \label{fig:SLAMMethods}
\end{figure}

Before presenting the results of the experiment on runtime monitors, it is essential to discuss the importance of choosing the appropriate SLAM methods. 
In this experiment, the actual speed is derived using SLAM's ability to publish the current position of the robot. However, there is a problem when using generic SLAM gmapping, as it relies on the wheel encoders of the robot. This means that any movement of the wheels is reflected in the SLAM's reported position, even if the physical robot itself hasn't actually moved. While this might not be an issue in ideal conditions, our experiment involves running the robot through rough terrain to examine the runtime monitor's ability to correct the difference between actual and expected speed, which can occur in such conditions. Therefore for this experiment, we are using the hector SLAM method. The advantage of using hector is obvious looking at Figure \ref{fig:SLAMMethods}. Unlike gmapping, which closely follows the wheel encoder data and can falsely indicate movement even when the robot is stationary (due to wheels slippage), hector SLAM relies solely on Lidar sensor data. This makes it more accurate in rough terrains, although it may still be susceptible to noise.

\begin{figure}
    \centering
    \includegraphics[scale=0.31]{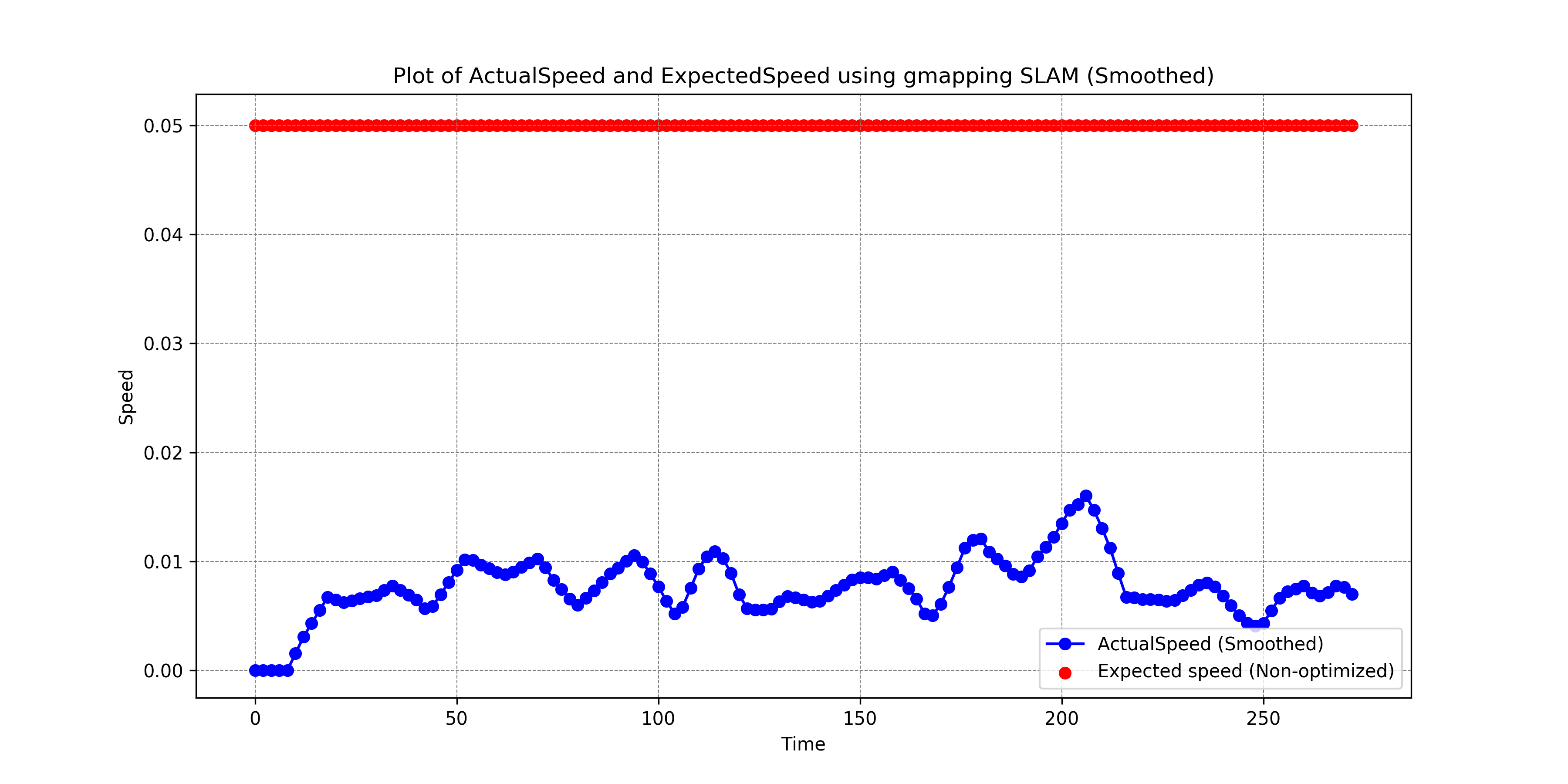}
\end{figure}

\subsection{Experiment Results}
T3B robot is run on a yoga mat with various things located underneath it, providing a bumpy low-density terrain.  The robot is programmed to test different linear speeds, ranging from 0.015 to 0.1. This helps in visualizing when the robot struggles to overcome various challenges in the terrain.

\begin{figure}
    \centering
    \includegraphics[scale=0.32]{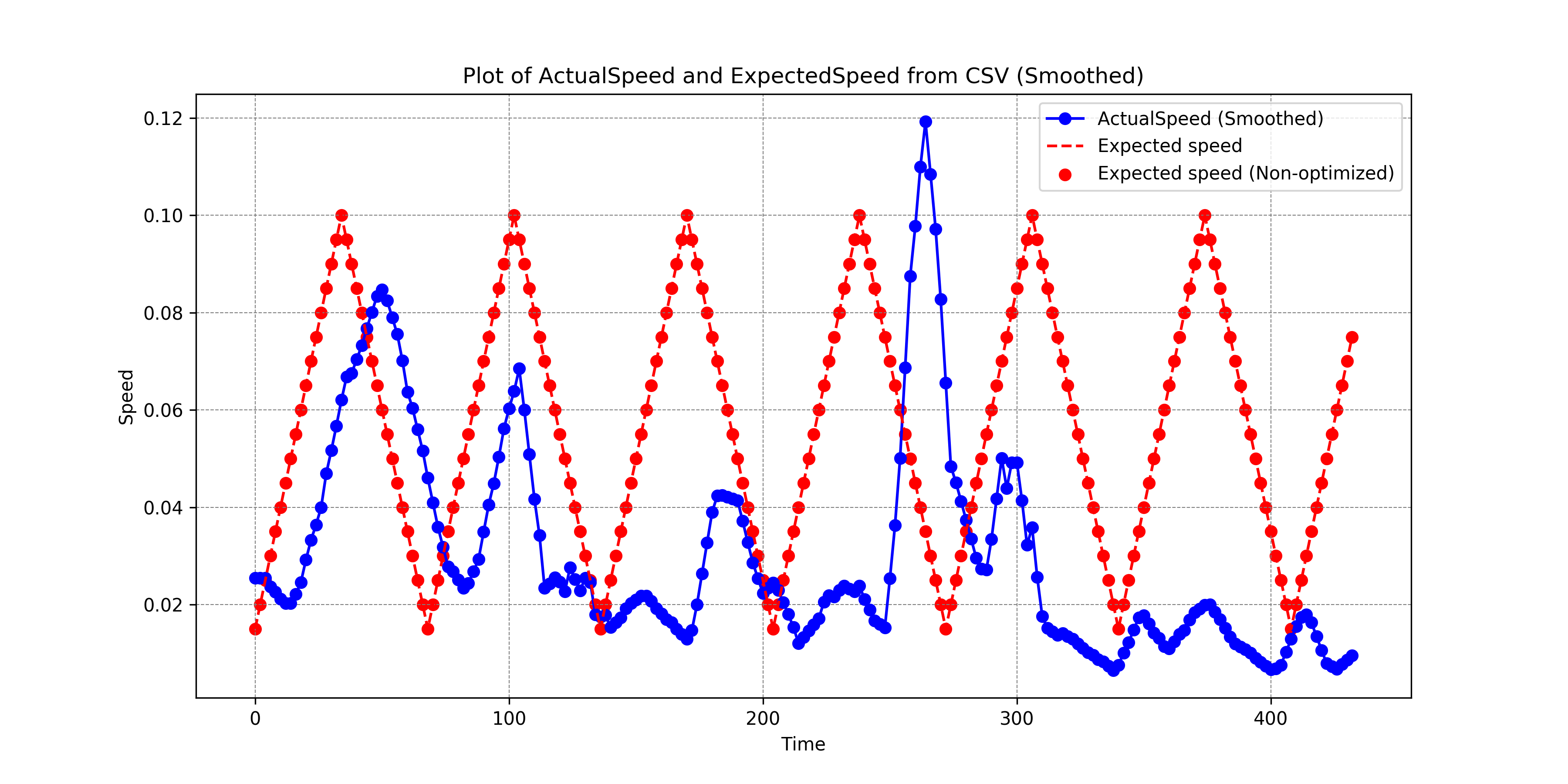}
    \caption{Plot of actual and expected speed using default robot navigation.}
    \label{fig:plotDefault}
\end{figure}

Figure \ref{fig:plotDefault} shows the actual and expected speeds when using the default robot navigation. As one can see, the robot achieves (almost) the expected speed, with a small delay, in the first 100 seconds. Afterward, it meets the first small hill in the terrain and is ultimately stuck until the expected speed reaches 0.1 again at around 250 seconds making the robot unstuck. Lastly, the robot gets stuck again and never exits before the execution is halted, making thus the difference between the actual and expected speed drastic.

\begin{figure}
    \centering
    \includegraphics[scale=0.32]{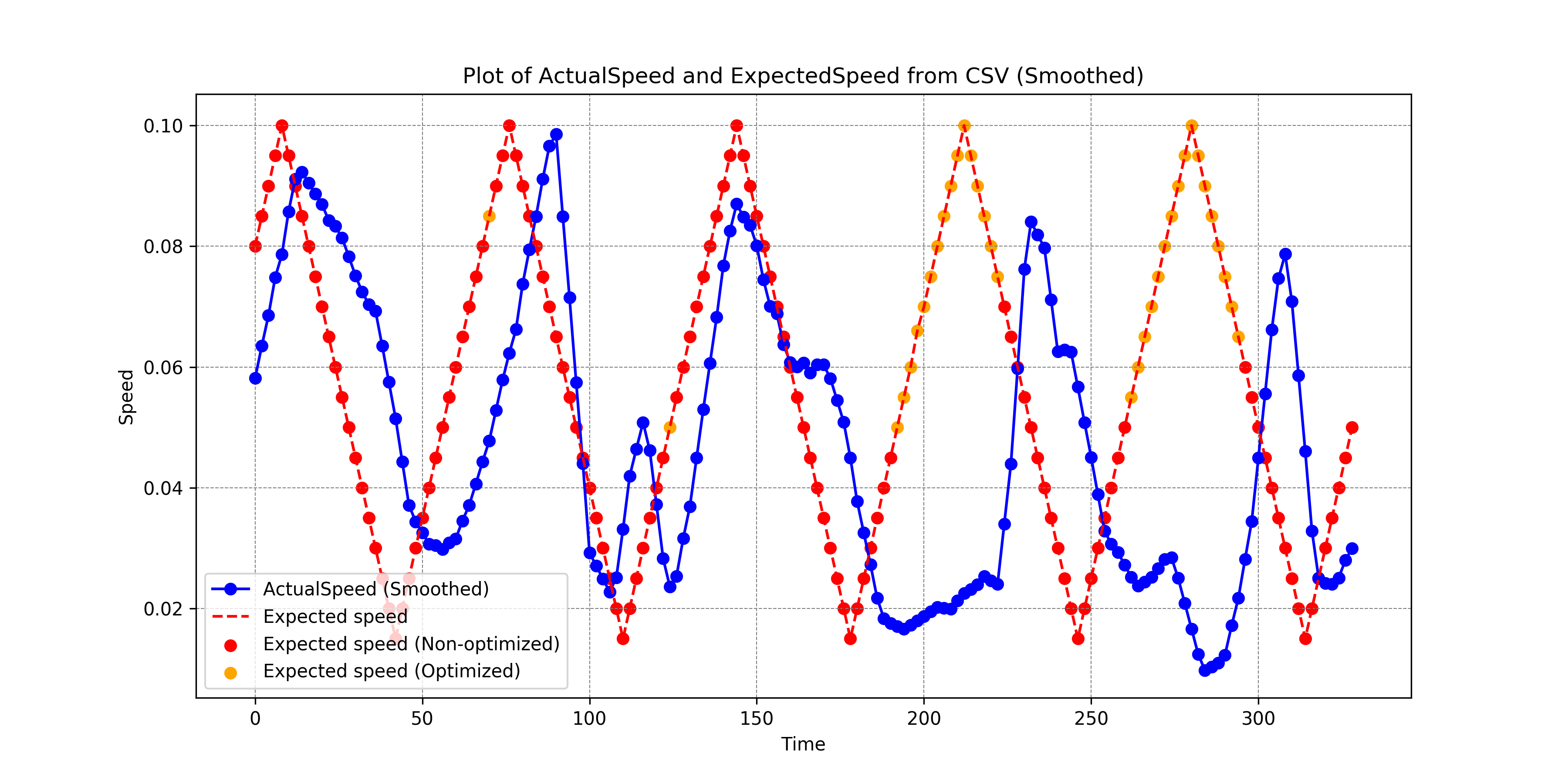}
    \caption{Plot of the actual and expected speeds when the robot is augmented with the DT runtime monitor for speed tolerance.}
    \label{fig:plotAugmented}
\end{figure}

Figure \ref{fig:plotAugmented} depicts the speed for the same experiment where the robot is augmented with the DT runtime monitors. Again, we observe that it drives smoothly until around 100 seconds when it meets the same small hill affecting the actual speed. The plot also shows where the runtime monitor is augmenting the expected speed of the robot to overcome these challenges showcased with the orange data points. We see one data point experiencing this at around 125, ultimately getting the robot over the small hill and achieving an actual speed close to the expected speed. The rest of the plot further demonstrates the robot encountering two additional challenges, affecting the actual speed, with the runtime monitor aiding the robot in overcoming these obstacles in the terrain. It can be observed that actual and expected speeds follow each other, albeit with a small delay, thanks to the effective intervention of the runtime monitor.

The comparison between the two figures clearly demonstrates the benefits of using the DT runtime monitor. With the default navigation, the robot struggles to maintain the expected speed over uneven terrain. In contrast, the DT runtime monitor significantly improves the robot's ability to handle challenging terrain by dynamically adjusting its speed.

To quantify the performance improvement, we calculate the Mean Squared Error (MSE) between the actual and expected speed which can be seen in Table \ref{tab:MSE}.

\begin{table}[ht]
\centering
\caption{MSE for Default Robot Navigation and Augmented Robot Navigation}
\label{tab:MSE}
\begin{tabular}{|l|c|}
\hline
\textbf{Navigation Mode}       & \textbf{MSE} \\ \hline
Default Robot Navigation       & 0.0017          \\ \hline
Augmented Robot Navigation     & 0.0010         \\ \hline
\end{tabular}
\end{table}

The numerical difference between 0.0017 and 0.0010 might seem minor, but the relative improvement of approximately 41\% indicates a significant enhancement in the robot's performance. The augmented navigation system demonstrates better speed tracking accuracy, increased robustness, and more reliable operation, particularly in challenging conditions. %These improvements can have a substantial impact on the overall effectiveness and efficiency of the robot's navigation capabilities.

\section{conclusion}
\label{sec:conclusion}
This paper proposed a digital twin empowered runtime verification framework for autonomous mobile robots, turtlebot 3 Burger, operating in uncertain environments. The uncertainty comes from the terrain conditions (density, elevation, etc) and faulty sensor data (due to noise, dust, occlusion, etc).  

The robot behavior constraints and safety properties are synthesized as runtime monitors in TeSSLA, implemented in Python and integrated in the (cloud-located) digital twin platform we designed for T3B robots. The synchronization of the executable digital twin, via MQTT protocol and Telegraf agent, with the robot enables continuous monitoring and validation of the robot’s behavior in real-time.

We have conducted different experiments to analyze the efficiency and time accuracy of the proposed runtime monitors using a physical robot and real-world scenarios. The experiment results demonstrate a high effectiveness of the proposed runtime monitoring and verification in ensuring the reliability and robustness of the autonomous robot behavior in uncertain environments. 

As a future work, we plan to augment the functionality of the runtime monitors to incorporate computation and coordination of the proper control actuations, and enhance the synchronization efficiency to reduce the time delays between the robot and its digital twin. 

%\section{Acknowledgment}

\bibliographystyle{abbrv}
\bibliography{main}

\end{document}